\documentclass[11pt]{article}

\usepackage[preprint]{acl}

\usepackage{times}
\usepackage{latexsym}
\usepackage[T1]{fontenc}
\usepackage[utf8]{inputenc}
\usepackage{microtype}
\usepackage{inconsolata}
\usepackage{graphicx}

\usepackage{amsmath,amsfonts,bm}

\def\eqref#1{equation~\ref{#1}}

\def\1{\bm{1}}

\DeclareMathAlphabet{\mathsfit}{\encodingdefault}{\sfdefault}{m}{sl}
\SetMathAlphabet{\mathsfit}{bold}{\encodingdefault}{\sfdefault}{bx}{n}

\usepackage{amsmath}
\usepackage{amssymb}
\usepackage{booktabs}
\usepackage[table]{xcolor}
\usepackage{subfig}
\usepackage{wrapfig}
\usepackage{multirow}
\usepackage{makecell}
\usepackage{tabularx}
\usepackage{array}
\usepackage{float}

\definecolor{ourscolor}{HTML}{f2f2fe}
\newcolumntype{C}{>{\centering\arraybackslash}X}

\title{PEAR: Adaptive Advantage Reshaping for Perception and Exploration in Large Vision-Language Models}

\author{
    Junxin Wang$^{1}$ \quad Yingfeng Luo$^{1}$ \quad Xiangchen Song$^{3}$ \quad Weiqiao Shan$^{1}$ \quad Runsong Zhao$^{1}$ \\
    \bfseries Xinyu Liu$^{1}$ \quad Shunjie Xing$^{1}$ \quad Yifu Huo$^{4}$ \quad Dingyang Lin$^{4}$ \quad Qiuyu Gu$^{1,2}$ \\
    \bfseries Yongyu Mu$^{3}$ \quad Chenglong Wang$^{1,2}$ \quad Changliang Li$^{1}$ \quad Jingbo Zhu$^{1,2}$ \quad Tong Xiao$^{1,2}$ \\[0.6em]
    \normalfont\small $^{1}$ School of Computer Science and Engineering, Northeastern University, Shenyang, China \\
    \normalfont\small $^{2}$ NiuTrans Research, Shenyang, China \\
    \normalfont\small $^{3}$ University of Michigan, Ann Arbor, US \\
    \normalfont\small $^{4}$ Kunming University of Science and Technology, Kunming, China
}

\begin{document}

\maketitle

\begin{abstract}
Effective multimodal reasoning requires a model to reliably utilize visual information while sufficiently exploring diverse reasoning paths. To achieve this goal, prior work combines visual dependency and token entropy to reshape sequence-level advantages, but scaling the contribution of token entropy by visual dependency can leave high-entropy tokens with low visual dependency underweighted, despite their important role in reasoning exploration. To address this limitation, we propose \textbf{PEAR}, an adaptive advantage reshaping method for improving perception and exploration in multimodal reasoning. PEAR retains visual dependency as the base signal and adds an entropy-based exploration gain only at high-entropy positions, with lower visual dependency leading to a larger additional gain. The fused signal is then mapped to positive token-level weights that reshape the sequence-level advantage. Experiments on Qwen2.5-VL-3B/7B show that PEAR achieves the highest average performance among evaluated baselines across multimodal reasoning benchmarks, alongside leading results on textual reasoning and general multimodal benchmarks.

\end{abstract}

\section{Introduction}
Reinforcement Learning with Verifiable Rewards (RLVR) has emerged as a promising approach to improving the reasoning capabilities of Large Language Models (LLMs)~\citep{deepseekr1,grpo}, but assigning credit across individual tokens within long reasoning trajectories remains challenging~\citep{grpo_lambda}. Recent work in text-only reasoning uses proxy signals such as token entropy to guide this assignment~\citep{erpo,wang2026beyond}, whereas multimodal reasoning requires both visual perception and the use of visual evidence throughout the reasoning process~\citep{tor}, making token-level credit assignment more challenging.

\begin{figure*}[t]
    \centering
    \includegraphics[width=\textwidth]{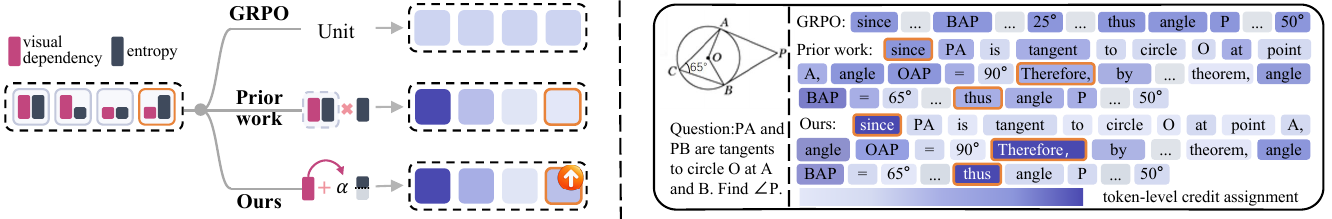}
    \caption{
    Comparison of token-level credit assignment strategies. \textbf{Left}: GRPO assigns the same credit to all tokens within a response, despite their different contributions to reasoning. Prior work fuses visual dependency and token entropy multiplicatively, limiting the influence of high entropy when visual dependency is low~\citep{pepo}. PEAR retains visual dependency as the basis for credit assignment and adaptively incorporates an exploration gain at high-entropy positions. \textbf{Right}: In the geometry example, PEAR gives greater credit to the reasoning transitions ``since'', ``Therefore'', and ``thus'' while continuing to emphasize vision-related tokens such as ``OAP'' and ``BAP''.
    }
    \label{fig:advantage_allocation}
\end{figure*}

To address this challenge, prior work jointly uses visual dependency and token entropy to construct differentiated token-level weights that reshape the shared sequence-level advantage~\citep{pepo}.
However, as illustrated in Figure~\ref{fig:advantage_allocation} (left), visual dependency scales the contribution of token entropy to these weights, so entropy has only limited influence on the weights of high-entropy tokens with low visual dependency, potentially leaving those tokens underweighted despite their role as critical decision points for exploring alternative reasoning paths~\citep{wang2026beyond}. For example, as shown in Figure~\ref{fig:advantage_allocation} (right), high-entropy transitions such as ``since,'' ``Therefore,'' and ``thus'' depend only weakly on visual information, yet they connect the steps through which intermediate conclusions are established and used to derive the final answer, and underweighting them can weaken the optimization signal at these important reasoning positions.

\begin{figure*}[t]
    \centering
    \includegraphics[width=\textwidth]{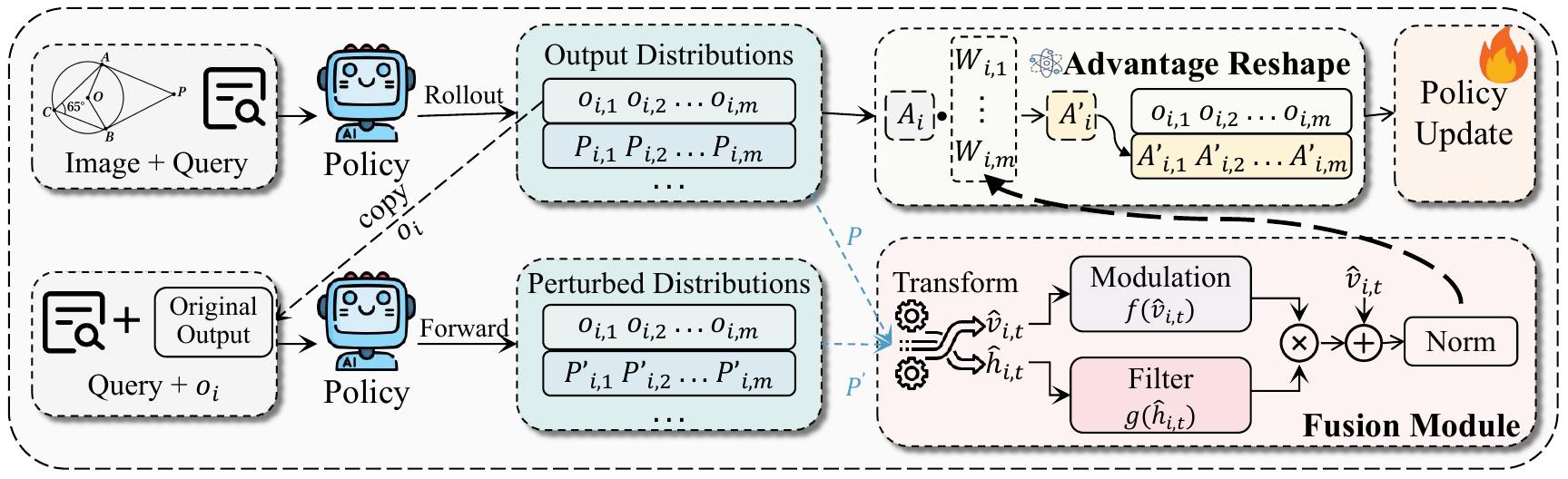}
    \caption{
    Overview of the proposed PEAR framework. Here, $o_{i,t}$ denotes the $t$-th token in response $i$, while $P_{i,t}$ and $P'_{i,t}$ denote the predictive distributions with and without image input, respectively. The Fusion Module derives visual dependency by comparing these distributions and estimates token entropy from $P_{i,t}$. The entropy signal is then filtered to produce exploration gains, which are adaptively modulated by visual dependency and added to the visual base to obtain normalized token weights. Advantage Reshape uses these weights to transform the shared sequence-level advantage into token-level advantages for policy update.
    }
    \label{fig:pear_framework}
\end{figure*}

To address this limitation, we propose \textbf{PEAR}, an adaptive advantage reshaping method for improving perception and exploration in multimodal reasoning. As illustrated in Figure~\ref{fig:pear_framework}, PEAR quantifies token-level visual dependency using the KL divergence between predictive distributions with and without image input and uses it as the base signal. Meanwhile, to account for the contribution of high-entropy tokens with low visual dependency, PEAR constructs exploration gains at high-entropy positions through a thresholded continuous entropy gate. Rather than adding these gains directly to the visual base, which would further amplify tokens already supported by strong visual signals, PEAR adaptively modulates the exploration gains before additive fusion, with lower visual dependency leading to a larger additional gain. The fused signal is then mapped to positive token-level weights, which are normalized within each response and used to reshape the sequence-level advantage while preserving its total mass.

We evaluate PEAR on multimodal reasoning benchmarks using Qwen2.5-VL-3B/7B. PEAR outperforms all evaluated baselines in average performance at both scales. Compared with prior work~\citep{pepo}, PEAR improves the 7B average by 2.61 points and outperforms it on every reported 3B benchmark. Notably, PEAR extends its strong performance to text-only reasoning and maintains its lead on general multimodal tasks. Ablation studies demonstrate the benefits of selectively and adaptively integrating visual dependency and token entropy.

\section{Preliminary}
\subsection{Group Relative Policy Optimization}
GRPO~\citep{grpo} performs policy optimization through group-wise relative evaluation without requiring an additional value model. Given a multimodal input \(x=(I,q)\), where \(I\) is the image and \(q\) is the textual question, it samples a group of \(G\) responses from the rollout policy \(\pi_{\theta_{\mathrm{old}}}\) and assigns a reward \(R_i\) to each response \(o_i\). By normalizing the group reward vector \(\mathbf{R}=(R_1,\ldots,R_G)\), GRPO obtains a sequence-level advantage that reflects each response's reward relative to the other responses generated for the same input:
\begin{equation}
\hat{A}_i = \frac{R_i-\operatorname{mean}(\mathbf{R})}{\operatorname{std}(\mathbf{R})}.
\end{equation}
Under outcome supervision, all tokens in \(o_i\) share the advantage \(\hat{A}_i\).
For token \(o_{i,t}\), the importance ratio conditioned on input \(x\) and prefix \(o_{i,<t}\) is
\begin{equation}
r_{i,t}(\theta) = \frac{\pi_{\theta}(o_{i,t} \mid x, o_{i,<t})}{\pi_{\theta_{\mathrm{old}}}(o_{i,t} \mid x, o_{i,<t})}.
\end{equation}
DAPO~\citep{dapo} extends GRPO with Clip-Higher and Token-Level Policy Gradient Loss. With \(r^{\mathrm{clip}}_{i,t}(\theta)=\operatorname{clip}\big(r_{i,t}(\theta),1-\epsilon_l,1+\epsilon_h\big)\), its policy-gradient objective is
\begin{equation}
\scalebox{0.91}{$\displaystyle
\begin{aligned}
\mathcal{J}_{\mathrm{DAPO}}(\theta)
&=\mathbb{E}\Bigl[
\frac{1}{\sum_{i=1}^{G}|o_i|}
\sum_{i=1}^{G}\sum_{t=1}^{|o_i|}\\
&\quad \min\Big(
r_{i,t}(\theta)\hat{A}_i,
r^{\mathrm{clip}}_{i,t}(\theta)\hat{A}_i
\Big)\Bigr].
\end{aligned}
$}
\end{equation}
Here, \(\epsilon_l\) and \(\epsilon_h\) specify the lower and upper clipping ranges, respectively.

\subsection{Visual Dependency and Token Entropy}

For the $t$-th token $o_{i,t}$ in response $o_i$, we characterize its visual dependency by comparing the probabilities assigned to the generated token with and without visual input. Given an image $I$, a textual question $q$, and the generated prefix $o_{i,<t}$, we follow PAPO~\citep{papo} and define the probability ratio between the conditions without and with visual input as 
\begin{equation}
r_{i,t}^{v}
=
\frac{
\pi_{\theta}(o_{i,t} \mid q,o_{i,<t})
}{
\pi_{\theta}(o_{i,t} \mid I,q,o_{i,<t})
}.
\end{equation}
The token-level visual dependency is then estimated using the low-variance KL estimator:
\begin{equation}
v_{i,t}
=
r_{i,t}^{v}
-
\log r_{i,t}^{v}
-
1.
\end{equation}
A larger $v_{i,t}$ indicates a greater change in the prediction of the current token when visual information is removed, suggesting stronger dependence on the visual input. To suppress extreme values, we further apply a logarithmic transformation, $\tilde{v}_{i,t}=\log(1+v_{i,t})$, which compresses large visual-dependency values while preserving their relative ordering across token positions.

For reasoning exploration, we use the generated token's negative log-probability under the original visual input as a single-sample Monte Carlo approximation of token entropy~\citep{papo}:
\begin{equation}
h_{i,t}
=
-\log
\pi_{\theta}(o_{i,t}\mid I,q,o_{i,<t}).
\end{equation}
This approximation avoids summing over the full vocabulary at each position. A larger $h_{i,t}$ corresponds to a lower generation probability and is used as a proxy for greater exploration potential at the current generation position.

\section{Method}
To enable fine-grained credit assignment, PEAR constructs token-level weights from visual dependency and token entropy to reshape the shared sequence-level advantage, assigning more credit to tokens with larger weights. These weights should reflect how strongly tokens rely on visual information while accounting for the role of high-entropy tokens with low visual dependency in reasoning exploration. To satisfy both requirements, PEAR normalizes the two signals, retains visual dependency as the base, and adaptively incorporates exploration gains at high-entropy positions, with the overall framework illustrated in Figure~\ref{fig:pear_framework}.

\subsection{Signal Normalization}

Visual dependency and token entropy have different numerical scales, so we apply Min-Max normalization to each signal within a response, obtaining $\hat{v}_{i,t}$ and $\hat{h}_{i,t}$. The normalized visual dependency is computed as
\begin{equation}
\hat{v}_{i,t}
=
\frac{
\tilde{v}_{i,t}-\min_j\tilde{v}_{i,j}
}{
\max_j\tilde{v}_{i,j}-\min_j\tilde{v}_{i,j}+\epsilon
}.
\end{equation}
Here, $j$ ranges over the token positions in response $o_i$.
The normalized entropy $\hat{h}_{i,t}$ is computed in the same way from $h_{i,t}$, where $\epsilon$ ensures numerical stability. This maps both signals to $[0,1]$ while preserving within-response ordering, providing comparable scales for fusion. Both normalized signals are detached and do not contribute gradients during policy optimization.

\subsection{Adaptive Signal Fusion}
PEAR aims to preserve the role of visual dependency in advantage assignment while accounting for the contribution of high-entropy tokens with low visual dependency. To this end, PEAR uses visual dependency as the base signal and introduces an exploration gain only at high-entropy positions, with this gain defined as follows:
\begin{equation}
e_{i,t}
=
\mathbf{1}\!\left[\hat{h}_{i,t}\geq\tau_h\right]
\left(
\frac{\hat{h}_{i,t}-\tau_h}
{1-\tau_h+\epsilon}
\right)^2,
\end{equation}
where $\mathbf{1}[\cdot]$ denotes the indicator function. When $\hat{h}_{i,t}<\tau_h$, we have $e_{i,t}=0$. Above the threshold, we normalize the entropy excess to prevent thresholding from making the gain too small, and then square it so that the gain grows slowly near the threshold but more rapidly at higher entropy levels. This design concentrates exploration gains on high-entropy positions while avoiding unnecessary modulation of low-entropy tokens.

Directly adding the entropy gain does not distinguish between high-entropy tokens with high and low visual dependency, assigning similar gains when their entropy values are comparable. Since the former already receive stronger weights from the visual base, such gains leave the within-sequence weight distribution biased toward them and limit the exploration contribution of the latter. PEAR therefore modulates the exploration gain according to visual dependency. The resulting fused signal is defined as:
\begin{equation}
c_{i,t}
=
\hat{v}_{i,t}
+
(1-\hat{v}_{i,t})e_{i,t}.
\end{equation}
When visual dependency is high, the fused signal remains primarily determined by the visual base, regardless of token entropy. Conversely, when visual dependency provides only a small base score and token entropy is high, the exploration gain can become the primary contributor to the fused signal.

\subsection{Advantage Reshaping}
Directly using the fused signal $c_{i,t}$ as the token weight may assign near-zero weights to positions with low fused scores, resulting in an excessively uneven weight distribution within a trajectory. To preserve inter-token differences while improving weight smoothness and training stability, we transform the fused signal into normalized token weights as follows:
\begin{equation}
\omega_{i,t}
=
\frac{
|o_i|\left(1+\beta c_{i,t}\right)
}{
\sum_{j=1}^{|o_i|}\left(1+\beta c_{i,j}\right)+\epsilon
}.
\end{equation}
Here, $|o_i|$ denotes the number of tokens in response $o_i$, and $\beta$ controls the modulation strength of the fused signal. The unit offset prevents token weights from approaching zero, while response-level normalization keeps their mean close to $1$. The resulting weights reflect the token-level differences in $c_{i,t}$ while controlling the overall scale of policy updates. The PEAR objective is then defined as:
\begin{equation}
\scalebox{0.91}{$\displaystyle
\begin{aligned}
\mathcal{J}_{\mathrm{PEAR}}(\theta)
&=\mathbb{E}\Bigl[
\frac{1}{\sum_{i=1}^{G}|o_i|}
\sum_{i=1}^{G}\sum_{t=1}^{|o_i|}\\
&\quad \boldsymbol{\omega}_{i,t}\min\Big(
r_{i,t}(\theta)\hat{A}_i,
r^{\mathrm{clip}}_{i,t}(\theta)\hat{A}_i
\Big)\Bigr].
\end{aligned}
$}
\end{equation}
By differentiating update strengths according to visual dependency and exploration potential, the token weights enable fine-grained credit assignment better suited to multimodal reasoning.

\begin{table*}[t]
    \centering
    \setlength{\tabcolsep}{4.2pt}
    \renewcommand{\arraystretch}{1.08}

    \resizebox{\textwidth}{!}{
    \begin{tabular}{l|cccccccc|c}
        \toprule
        \multicolumn{1}{c|}{\textbf{Model}}
        & \textbf{Geo3K}
        & \textbf{DynaMath}
        & \textbf{We-Math}
        & \textbf{MathVision}
        & \textbf{MathVerse}
        & \textbf{LogicVista}
        & \textbf{MMMU-Pro}
        & \textbf{MathVerse$_V$}
        & \textbf{Avg.} \\
        \midrule

        MM-Eureka-7B
        & 40.77
        & 49.53
        & 63.78
        & \underline{28.49}
        & 66.44
        & \underline{46.92}
        & 23.42
        & 61.76
        & 47.64 \\

        VL-Rethinker-7B
        & 39.45
        & 52.38
        & 67.96
        & 27.91
        & 67.86
        & 45.41
        & \textbf{37.05}
        & 64.58
        & 50.33 \\

        NoisyRollout-7B
        & \textbf{52.12}
        & 53.65
        & 63.27
        & 27.95
        & 55.97
        & 46.47
        & 33.49
        & 60.04
        & 49.12 \\

        R1-ShareVL-7B
        & 42.12
        & 53.27
        & \underline{69.48}
        & 27.95
        & 67.40
        & 46.64
        & 35.01
        & 63.69
        & 50.70 \\

        Vision-SR1-7B
        & 36.71
        & 51.45
        & 63.05
        & 26.62
        & 60.33
        & 44.85
        & 35.63
        & 56.06
        & 46.84 \\

        \midrule

        \textit{Qwen2.5-VL-7B}
        & 37.52
        & 48.06
        & 46.08
        & 25.25
        & 39.21
        & 41.30
        & 25.56
        & 34.50
        & 37.19 \\

        \multicolumn{1}{r|}{+ GRPO}
        & 42.24
        & 54.84
        & 68.15
        & 27.66
        & 67.91
        & 45.50
        & 34.89
        & 63.57
        & 50.60 \\

        \multicolumn{1}{r|}{+ DAPO}
        & 35.48
        & \underline{55.96}
        & 63.77
        & 28.16
        & 60.92
        & 41.08
        & 31.60
        & 57.22
        & 46.77 \\

        \multicolumn{1}{r|}{+ PAPO}
        & 43.78
        & 55.89
        & 68.03
        & 28.28
        & \underline{68.75}
        & 46.45
        & 36.42
        & \underline{65.13}
        & \underline{51.59} \\

        \multicolumn{1}{r|}{+ PEPO}
        & 39.62
        & 54.49
        & 68.82
        & 28.01
        & 66.11
        & 46.20
        & 35.33
        & 62.17
        & 50.09 \\

        \rowcolor{blue!6}
        \multicolumn{1}{r|}{+ PEAR}
        & \underline{45.13}
        & \textbf{56.15}
        & \textbf{69.65}
        & \textbf{29.40}
        & \textbf{70.28}
        & \textbf{47.96}
        & \underline{36.55}
        & \textbf{66.50}
        & \textbf{52.70} \\

        \midrule
        \midrule

        \textit{Qwen2.5-VL-3B}
        & 19.01
        & 33.12
        & 30.42
        & 17.24
        & 33.20
        & 30.45
        & 19.36
        & 30.11
        & 26.61 \\

        \multicolumn{1}{r|}{+ GRPO}
        & 29.97
        & 44.21
        & 59.06
        & 23.39
        & 56.10
        & 38.76
        & 26.16
        & 52.90
        & 41.32 \\

        \multicolumn{1}{r|}{+ DAPO}
        & 33.47
        & 45.86
        & 62.79
        & \underline{26.19}
        & 59.56
        & 41.89
        & \underline{28.28}
        & 56.72
        & 44.35 \\

        \multicolumn{1}{r|}{+ PAPO}
        & \textbf{36.67}
        & \underline{47.30}
        & \underline{63.46}
        & 26.16
        & \underline{59.97}
        & \underline{42.25}
        & 27.93
        & \underline{57.07}
        & \underline{45.10} \\

        \multicolumn{1}{r|}{+ PEPO}
        & 29.85
        & -
        & 60.98
        & -
        & 56.49
        & 39.85
        & 27.47
        & -
        & - \\

        \rowcolor{blue!6}
        \multicolumn{1}{r|}{+ PEAR}
        & \underline{36.13}
        & \textbf{48.94}
        & \textbf{63.65}
        & \textbf{26.71}
        & \textbf{62.28}
        & \textbf{43.99}
        & \textbf{28.80}
        & \textbf{59.48}
        & \textbf{46.25} \\

        \bottomrule
    \end{tabular}
    }
    \caption{
        Performance comparison on eight multimodal reasoning benchmarks using the Avg@8 metric.
        The best and second-best results within each model scale are
        highlighted in bold and underlined, respectively.
    }
    \label{tab:main_results}
\end{table*}

\begin{table*}[t]
    \centering
    \setlength{\tabcolsep}{6.0pt}
    \renewcommand{\arraystretch}{1.08}

    \resizebox{\textwidth}{!}{
    \begin{tabular}{l|ccccc|ccccc}
        \toprule
        \multirow{2}{*}{\textbf{Model}}
        & \multicolumn{5}{c|}{\textbf{Textual Reasoning}}
        & \multicolumn{5}{c}{\textbf{General Multimodal}} \\
        \cmidrule(lr){2-6}
        \cmidrule(lr){7-11}

        & \textbf{AIME24}
        & \textbf{MATH-500}
        & \textbf{BBH}
        & \textbf{MMLU-Pro}
        & \textbf{Avg.}
        & \textbf{MMStar}
        & \textbf{MMMU}
        & \textbf{MME (\%)}
        & \textbf{POPE(F1)}
        & \textbf{Avg.} \\
        \midrule

        MM-Eureka-7B
        & \underline{6.46}
        & \textbf{64.00}
        & 61.99
        & 52.01
        & \underline{46.12}
        & 57.70
        & 56.11
        & 83.36
        & 85.80
        & 70.74 \\

        VL-Rethinker-7B
        & 4.38
        & 60.40
        & 62.99
        & 52.40
        & 45.04
        & 60.90
        & 56.55
        & 83.06
        & 85.80
        & 71.58 \\

        NoisyRollout-7B
        & 6.04
        & \underline{61.40}
        & 59.13
        & \underline{52.58}
        & 44.79
        & 61.50
        & 56.00
        & 83.12
        & 86.50
        & 71.78 \\

        R1-ShareVL-7B
        & 3.75
        & 56.40
        & 64.21
        & 52.41
        & 44.19
        & 61.30
        & 55.11
        & 83.46
        & 86.60
        & 71.62 \\

        Vision-SR1-7B
        & 5.00
        & 56.80
        & 62.92
        & 52.19
        & 44.23
        & 33.70
        & 55.33
        & 82.70
        & 85.90
        & 64.41 \\

        \midrule

        \textit{Qwen2.5-VL-7B}
        & 4.79
        & 58.00
        & 53.86
        & 50.22
        & 41.72
        & 60.80
        & 52.56
        & 81.58
        & 85.82
        & 70.19 \\

        \multicolumn{1}{r|}{+ GRPO}
        & 6.25
        & 58.80
        & 63.19
        & 52.28
        & 45.13
        & \textbf{62.10}
        & 55.22
        & 83.63
        & \textbf{86.80}
        & 71.94 \\

        \multicolumn{1}{r|}{+ DAPO}
        & 2.08
        & 49.60
        & 63.52
        & 50.49
        & 41.42
        & \textbf{62.10}
        & 56.11
        & 83.45
        & 84.80
        & 71.61 \\

        \multicolumn{1}{r|}{+ PAPO}
        & 3.54
        & 58.80
        & \textbf{64.92}
        & 52.29
        & 44.89
        & 61.30
        & 56.33
        & \textbf{83.93}
        & \underline{86.70}
        & 72.06 \\

        \multicolumn{1}{r|}{+ PEPO}
        & 4.79
        & 59.00
        & 63.94
        & 52.44
        & 45.04
        & \underline{61.80}
        & \underline{56.56}
        & 83.48
        & 86.52
        & \underline{72.09} \\

        \rowcolor{blue!6}
        \multicolumn{1}{r|}{+ PEAR}
        & \textbf{7.29}
        & 59.40
        & \underline{64.68}
        & \textbf{53.46}
        & \textbf{46.21}
        & \textbf{62.10}
        & \textbf{57.67}
        & \underline{83.90}
        & 86.10
        & \textbf{72.44} \\

        \midrule
        \midrule

        \textit{Qwen2.5-VL-3B}
        & 3.54
        & 51.20
        & 48.52
        & 41.17
        & 36.11
        & 54.40
        & 41.44
        & 78.24
        & 85.67
        & 64.94 \\

        \multicolumn{1}{r|}{+ GRPO}
        & 3.54
        & 52.00
        & 54.00
        & 43.53
        & 38.27
        & 55.87
        & 44.11
        & 78.36
        & \underline{86.57}
        & \underline{66.23} \\

        \multicolumn{1}{r|}{+ DAPO}
        & \underline{3.75}
        & \underline{59.00}
        & \underline{55.66}
        & \underline{45.89}
        & \underline{41.08}
        & 55.73
        & 43.78
        & 77.58
        & \textbf{87.27}
        & 66.09 \\

        \multicolumn{1}{r|}{+ PAPO}
        & 3.33
        & 56.60
        & 55.18
        & 45.44
        & 40.14
        & \underline{56.33}
        & \underline{44.44}
        & \textbf{78.50}
        & 85.40
        & 66.17 \\

        \rowcolor{blue!6}
        \multicolumn{1}{r|}{+ PEAR}
        & \textbf{5.62}
        & \textbf{60.00}
        & \textbf{56.12}
        & \textbf{46.17}
        & \textbf{41.98}
        & \textbf{56.60}
        & \textbf{46.33}
        & \underline{78.48}
        & 86.10
        & \textbf{66.88} \\

        \bottomrule
    \end{tabular}
    }
    \caption{
    Performance comparison on textual reasoning and general multimodal ability benchmarks.
    The best and second-best results within each model scale are highlighted in bold and underlined, respectively.
    MME scores are normalized by the maximum score of 2800 and reported as percentages.
    PEPO-3B results on these benchmarks are not reported in the original work~\citep{pepo} and are therefore not included in this table.
    }
    \label{tab:general_capability}
\end{table*}

\section{Experiments}

\subsection{Experimental Setup}
\paragraph{Models, Data, and Baselines.}
We train Qwen2.5-VL-3B/7B-Instruct~\citep{qwen2} on ViRL39K~\citep{vl_rethinker}, which contains only questions and final answers. The primary baselines are GRPO~\citep{grpo}, DAPO~\citep{dapo}, PAPO~\citep{papo}, and PEPO~\citep{pepo}. GRPO and DAPO use the same initial checkpoints and training data as PEAR, whereas PAPO is evaluated using its publicly released checkpoints under our evaluation protocol. Other 7B baselines include MM-Eureka-7B~\citep{mm_eureka}, VL-Rethinker-7B~\citep{vl_rethinker}, NoisyRollout-7B~\citep{noisyrollout}, R1-ShareVL-7B~\citep{sharevl}, and Vision-SR1-7B~\citep{Vision-SR1}. For PEPO, we use its reported 3B results on ViRL39K and train the 7B model using its publicly released code.
\paragraph{Training Details.}
Our models are trained on ViRL39K for 2 epochs with a rollout batch size of 384, a maximum response length of 2048, and 5 responses sampled per input. The rule-based reward combines answer accuracy with format compliance. Following DAPO~\citep{dapo}, we use asymmetric clipping with \(\epsilon_l=0.2\) and \(\epsilon_h=0.28\), together with token-level loss normalization and dynamic sampling based on answer accuracy. We set the high-entropy threshold to \(\tau_h=0.5\) and the advantage modulation coefficient to \(\beta=1.2\). For Qwen2.5-VL-7B-Instruct, we add an entropy penalty with coefficient \(0.03\) to stabilize training, as in prior multimodal RL work~\citep{papo}, and anneal the exploration-gain coefficient with a cosine schedule (see Appendix~\ref{app:stabilization_ablation} for ablations of stabilization designs). For all models trained in this work, we report results from the final checkpoint.
\paragraph{Evaluation Benchmarks.}
We evaluate \textbf{Multimodal Reasoning} performance on eight benchmarks: Geo3K~\citep{geo3k}, DynaMath~\citep{dynamath}, We-Math~\citep{we_math}, MathVision~\citep{mathvision}, MathVerse and its vision-centric subset MathVerse$_V$~\citep{mathverse}, LogicVista~\citep{logicvista}, and MMMU-Pro~\citep{mmmu_pro}.
For generalization, we use two further categories of benchmarks: \textbf{Textual Reasoning}, including AIME24~\citep{aime24}, MATH-500~\citep{math500}, \mbox{BBH~\citep{bbh}}, and MMLU-Pro~\citep{mmlu_pro}; and \textbf{General Multimodal}, including MMStar~\citep{mmstar}, MMMU~\citep{mmmu}, MME~\citep{mme}, and POPE~\citep{pope}.
For the multimodal reasoning benchmarks, we report Avg@8. For the generalization evaluation, we report Avg@1 on all benchmarks except AIME24, for which Avg@16 is reported. More details on the benchmarks and evaluation protocols are provided in Appendix~\ref{app:benchmark_details}.

\subsection{Results}
\paragraph{Multimodal Reasoning.}
As shown in Table~\ref{tab:main_results}, PEAR achieves the highest average scores among evaluated methods at both model scales. Under the same training settings, PEAR maintains a clear advantage over GRPO, achieving relative improvements in average performance of approximately \textbf{11.9\%} (\textbf{41.32 $\rightarrow$ 46.25}) and \textbf{4.2\%} (\textbf{50.60 $\rightarrow$ 52.70}) on the 3B and 7B models, respectively, while achieving improvements across all eight evaluated benchmarks at both model scales. Compared with PEPO, which also reshapes token-level advantages using visual and entropy signals, PEAR improves the 7B average by \textbf{2.61} points (\textbf{50.09 $\rightarrow$ 52.70}), with higher scores on every benchmark. For the 3B model, PEAR achieves gains of up to \textbf{6.28} points, with at least a \textbf{1.33}-point improvement on every benchmark reported by PEPO.

\paragraph{Textual Reasoning.}
Beyond multimodal reasoning, Table~\ref{tab:general_capability} shows that PEAR consistently outperforms GRPO across all four textual benchmarks at both model scales. On Qwen2.5-VL-3B, PEAR achieves an average score of 41.98, outperforming GRPO by 3.71 points and yielding the best overall performance among the compared 3B methods. On Qwen2.5-VL-7B, PEAR achieves the highest average score of 46.21, surpassing GRPO by 1.08 points. Although PEPO also incorporates entropy, its entropy-driven modulation remains constrained by visual dependency, limiting the reinforcement of reasoning tokens, whereas PEAR provides additional gains to high-entropy positions with low visual dependency and achieves a 1.17-point higher average score on text-only reasoning benchmarks.
\paragraph{General Multimodal.}
Although PEAR is primarily designed to improve multimodal reasoning, the results in Table~\ref{tab:general_capability} show that it also achieves leading average performance among the compared RL methods on general multimodal benchmarks, scoring 66.88 and 72.44 on Qwen2.5-VL-3B and Qwen2.5-VL-7B, respectively. Its gains are particularly pronounced on MMMU, where it outperforms GRPO by 2.22 and 2.45 points on the 3B and 7B models, respectively. Without dedicated training on these tasks, PEAR retains general multimodal capability and achieves modest average gains on both the 3B and 7B models.

\begin{table*}[t]
    \centering
    \setlength{\tabcolsep}{4.2pt}
    \renewcommand{\arraystretch}{1.08}

    \resizebox{\textwidth}{!}{
    \begin{tabular}{l|cccccccc|c}
        \toprule
        \multicolumn{1}{c|}{\textbf{Setting}}
        & \textbf{Geo3K}
        & \textbf{DynaMath}
        & \textbf{We-Math}
        & \textbf{MathVision}
        & \textbf{MathVerse}
        & \textbf{LogicVista}
        & \textbf{MMMU-Pro}
        & \textbf{MathVerse$_V$}
        & \textbf{Avg.} \\
        \midrule

        GRPO
        & 29.97
        & 44.21
        & 59.06
        & 23.39
        & 56.10
        & 38.76
        & 26.16
        & 52.90
        & 41.32 \\

        \midrule
        \multicolumn{10}{c}{
            \textit{Individual Signals}
        } \\
        \midrule

        Visual Base Only
        & 33.46
        & 46.83
        & 58.69
        & 25.51
        & 54.40
        & 41.08
        & 28.11
        & 52.05
        & 42.52 \\

        High-Entropy Gain Only
        & 34.94
        & 47.80
        & 63.74
        & 25.46
        & 60.36
        & 41.92
        & 27.69
        & 57.23
        & 44.89 \\

        \midrule
        \multicolumn{10}{c}{
            \textit{Component Ablation}
        } \\
        \midrule

        w/o Visual Modulation
        & 36.29
        & 47.93
        & \textbf{64.35}
        & 26.36
        & 61.26
        & 41.81
        & 27.78
        & 58.57
        & 45.54 \\

        w/o Entropy Filtering
        & \underline{36.92}
        & \underline{48.79}
        & 63.23
        & \textbf{26.97}
        & \underline{61.97}
        & \underline{42.23}
        & 28.33
        & 58.24
        & \underline{45.84} \\

        \midrule
        \multicolumn{10}{c}{
            \textit{Entropy Threshold ($\tau_h$)}
        } \\
        \midrule

        $\tau_h = 0.25$
        & 29.92
        & 44.59
        & 58.97
        & 23.51
        & 55.95
        & 38.61
        & 27.00
        & 52.10
        & 41.33 \\

        $\tau_h = 0.75$
        & 34.81
        & 46.76
        & 62.90
        & 24.62
        & 60.71
        & 41.77
        & 27.77
        & 57.13
        & 44.56 \\

        \midrule
        \multicolumn{10}{c}{
            \textit{Advantage Modulation Strength ($\beta$)}
        } \\
        \midrule

        $\beta = 1.0$
        & 35.00
        & 48.02
        & 62.79
        & 26.10
        & 59.41
        & 41.92
        & \underline{28.72}
        & 55.73
        & 44.71 \\

        $\beta = 1.4$
        & \textbf{37.33}
        & 48.02
        & \underline{64.14}
        & 26.34
        & 61.53
        & 41.61
        & 28.33
        & \underline{58.72}
        & 45.75 \\

        \midrule
        \rowcolor{blue!6}
        PEAR
        & 36.13
        & \textbf{48.94}
        & 63.65
        & \underline{26.71}
        & \textbf{62.28}
        & \textbf{43.99}
        & \textbf{28.80}
        & \textbf{59.48}
        & \textbf{46.25} \\

        \bottomrule
    \end{tabular}
    }
    \caption{
        Ablation studies of PEAR on Qwen2.5-VL-3B.
        All experiments are conducted under the same training setting
        unless otherwise specified.
        The best and second-best results among all evaluated settings are highlighted in bold and underlined, respectively.
        For each hyperparameter sweep, the other hyperparameter is fixed at its default value, with $\tau_h=0.5$ and $\beta=1.2$.
    }
    \label{tab:ablation}
\end{table*}

\section{Analysis}
\subsection{Token Analysis}
\begin{figure}[H]
    \centering
    \includegraphics[width=\columnwidth]{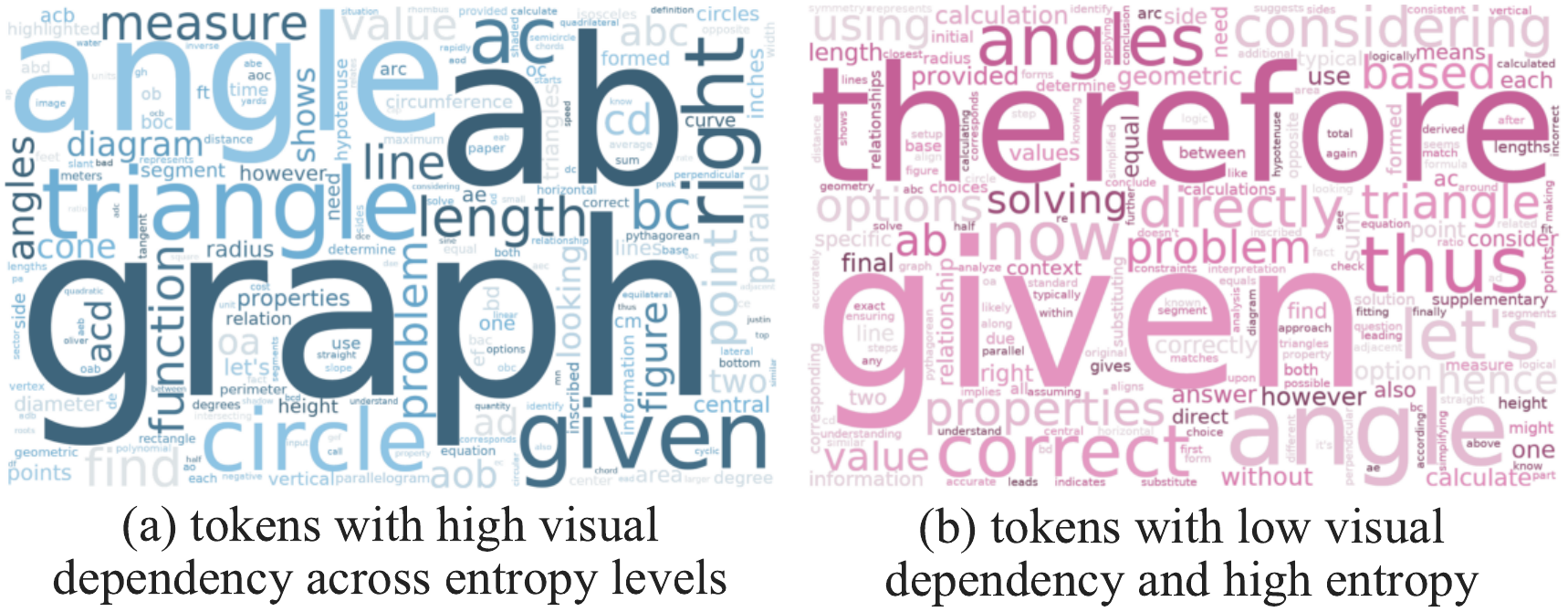}
    \caption{Token word clouds on MathVerse: (a) high visual dependency; (b) low visual dependency and high entropy. Font size indicates frequency.}
    \label{fig:wordcloud}
\end{figure}

To examine which types of tokens receive stronger optimization signals under PEAR, we construct word clouds for tokens with high visual dependency across entropy levels and those with low visual dependency but high entropy, using correct reasoning trajectories generated by Qwen2.5-VL-3B-Instruct on MathVerse. As shown in Fig.~\ref{fig:wordcloud}, tokens with high visual dependency are dominated by terms describing visual entities and geometric attributes, whereas the group with low visual dependency and high token entropy contains more expressions associated with reasoning transitions and intermediate judgments. This contrast suggests that the latter may carry information related to reasoning that complements visually grounded content, supporting the motivation behind PEAR. Construction details are provided in Appendix~\ref{app:wordcloud_construction}.

\subsection{Training Dynamics}
\begin{figure}[!t]
    \centering
    \includegraphics[width=\columnwidth,trim=2pt 2pt 3pt 2pt,clip]{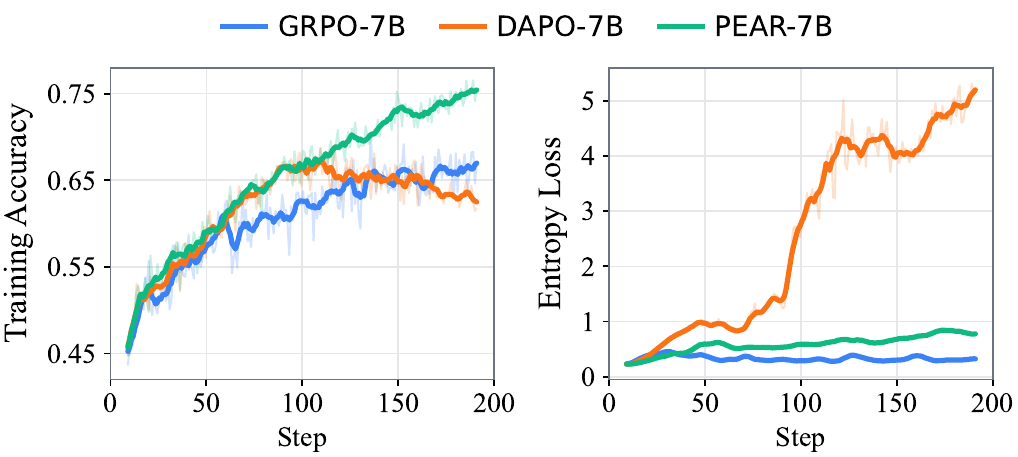}
    \caption{Training dynamics of Qwen2.5-VL-7B during policy optimization. PEAR maintains its exploration capability while continuing to improve training accuracy.}
    \label{fig:training_dynamics}
\end{figure}

The effectiveness of PEAR is underpinned by its superior training dynamics, as illustrated by the training curves against the baselines in Fig.~\ref{fig:training_dynamics}. Compared with GRPO, PEAR maintains an overall higher token entropy estimate and continues to improve its training accuracy after GRPO gradually plateaus; as training proceeds, DAPO's entropy estimate rises rapidly and is accompanied by a decline in training accuracy, whereas PEAR maintains stable trajectories in both entropy and training accuracy. We attribute these training dynamics largely to the fact that PEAR does not directly maximize entropy, but instead uses token entropy to identify positions with greater exploration potential, thereby maintaining the model's exploration capability during policy optimization.

\subsection{Ablation}

\paragraph{Individual Signals.}
To further examine the individual roles of the visual base and high-entropy gain, we construct two variants that retain one of the signals. As shown in Table~\ref{tab:ablation}, Visual Base Only achieves an average score of 42.52, below the 46.25 achieved by full PEAR. We conjecture that this reflects a skewed distribution of visual dependency: only a small subset of tokens has strong visual signals, while most have weak ones. Using visual dependency alone for weighting may thus be susceptible to noise and require a tailored mapping or smoothing function to maintain performance. The High-Entropy Gain Only variant directs additional optimization toward uncertain reasoning positions and remains stable during training, but achieves an average score of 44.89, which is 1.36 points below the 46.25 achieved by the full PEAR. Combining the two signals allows PEAR to maintain stable training while improving model performance.

\paragraph{Component.}
To evaluate the roles of entropy filtering and visual dependency modulation, we remove each component separately. Removing visual dependency modulation assigns similar exploration gains to high-entropy tokens with similar entropy values regardless of their visual dependency, so those with lower visual dependency no longer receive stronger supplementation. The average performance drops from 46.25 to 45.54. Without entropy filtering, exploration gains also extend to low-entropy tokens, weakening the focus of advantage reshaping on critical reasoning positions and lowering the average performance to 45.84. These drops support retaining both components in PEAR.
\paragraph{Hyperparameters.}
We analyze the sensitivity of PEAR to two key hyperparameters: the entropy threshold $\tau_h$ and the advantage modulation strength $\beta$. As shown in Table~\ref{tab:ablation}, PEAR achieves the best average performance among the evaluated settings at $\tau_h=0.5$ and $\beta=1.2$. For $\tau_h$, lowering the threshold allows more low-entropy tokens to participate in advantage reshaping, potentially diluting the focus on uncertain reasoning positions. At $\tau_h=0.25$, the average score drops to 41.33 from 46.25. In contrast, an overly high threshold makes the selection too restrictive, leaving fewer uncertain reasoning tokens to receive additional emphasis and thereby limiting exploration. For $\beta$, a small value provides insufficient modulation of the selected tokens, limiting the effect of advantage reshaping, whereas a larger value may overemphasize token-level differences within a trajectory and reduce performance.

\subsection{Computational Overhead}
As shown in Table~\ref{tab:training_efficiency}, PEAR achieves noticeable performance gains while introducing only moderate additional training overhead. The extra training time of PEAR mainly comes from the image-free forward pass required to compute visual dependency, whereas the additional costs introduced by entropy computation, information fusion, and subsequent advantage reshaping are almost negligible. PEAR also follows the dynamic sampling strategy of DAPO, yet still incurs lower overall training cost than vanilla DAPO. We speculate that this may be related to PEAR maintaining more sufficient generation exploration during training, which leads to more diverse responses for the same input and makes it easier to form sample groups with effective reward differences, thereby reducing the extra resampling cost caused by filtering invalid groups during dynamic sampling.
\begin{table}[H]
    \centering
    \small
    \setlength{\tabcolsep}{5pt}
    \renewcommand{\arraystretch}{1.1}

    \resizebox{\columnwidth}{!}{%
    \begin{tabular}{lcccc}
        \toprule

        \multirow{2}{*}{Method}
        &
        \multirow{2}{*}{Time (h)}
        &
        Multimodal
        &
        Textual
        &
        General \\

        &
        &
        Reasoning
        &
        Reasoning
        &
        Multimodal \\

        \midrule

        GRPO
        & 14.9 (+0.0\%)
        & 50.60
        & 45.13
        & 71.94 \\

        DAPO
        & 21.5 (+44.3\%)
        & 46.77
        & 41.42
        & 71.61 \\

        \rowcolor{gray!15}
        PEAR
        & 18.1 (+21.5\%)
        & \textbf{52.70}
        & \textbf{46.21}
        & \textbf{72.44} \\

        \bottomrule
    \end{tabular}
    }
    \caption{Training time on eight NVIDIA A100 GPUs and average performance of Qwen2.5-VL-7B under the same experimental setup.}
    \label{tab:training_efficiency}
\end{table}

\section{Related Work}
Existing studies on multimodal reasoning mainly focus on improving reasoning capabilities in complex visual tasks.
These approaches broadly fall into three directions: visual perception, reasoning processes, and their joint optimization.

\paragraph{Visual Perception.}
Methods centered on perception improve visual information utilization through perceptual supervision, explicit grounding, and perception-guided optimization.
Perceptual supervision uses verifiable visual rewards or auxiliary perception objectives to improve visual understanding and grounding~\citep{visualrft,vicrit,perceptionr1}.
Explicit grounding methods incorporate visual information into the reasoning process through visual evidence, grounded reasoning trajectories, or intermediate representations that encode visual information~\citep{pointrft,vigorl,rapid}. 
Perception-guided optimization methods use visual dependency or relevance signals to optimize multimodal reasoning models~\citep{vppo,vgpo,pgpo,li2026prpo}.

\paragraph{Reasoning.}
Methods centered on reasoning enhance multimodal reasoning through explicit reasoning supervision, reinforcement learning, and reasoning process refinement.
Explicit reasoning supervision methods introduce chain of thought (CoT) trajectories to enable models to learn sequential problem solving processes and improve complex reasoning capabilities~\citep{llava_cot}. 
Following the success of large language models designed for reasoning, reinforcement learning has become an important approach for optimizing multimodal reasoning processes, where verifiable rewards are utilized to encourage the exploration of more effective reasoning trajectories and improve reasoning over extended horizons~\citep{mm_eureka,lmm_r1,vlm_r1,visual_r1}. 
Moreover, some studies improve reasoning reliability through reflection, iterative refinement, and consistency-guided optimization~\citep{vl_rethinker,openvlthinker}.

\paragraph{Joint Perception and Reasoning Optimization.}
Joint methods coordinate perception and reasoning through the optimization objective, reward design, or token-level credit assignment. At the objective level, PAPO introduces an implicit perception loss to strengthen visual grounding during reasoning optimization~\citep{papo}. At the reward level, PEARL anchors reasoning updates to verified perceptual evidence, while PRCO assigns rewards tailored to cooperative observer and solver roles~\citep{pearl,prco}. At the token level, ToR and SRPO distinguish perception and reasoning tokens through dynamic reweighting and role-specific credit assignment, respectively~\citep{tor,srpo}. VEPO instead selects informative tokens by jointly considering visual sensitivity and token entropy~\citep{vepo}. Moving beyond token identification and selection, PEPO introduces an advantage reshaping method at the token level that jointly models a perception prior derived from visual similarity and token entropy to transform advantages defined at the sequence level into differentiated optimization signals for individual tokens, yet it cannot fully preserve the contribution of token entropy to reasoning exploration~\citep{pepo}. Concurrently, TPAE estimates token-level advantages by measuring how each token's visual dependency and predictive entropy align with reference distributions from correct rollouts to identify outliers~\citep{zhang2026tpae}.

\section{Conclusion}
In this work, we introduce PEAR, an adaptive advantage reshaping method for improving perception and exploration in multimodal reasoning. PEAR retains visual dependency as the base signal and adds an entropy-based exploration gain only at high-entropy positions, with lower visual dependency leading to a larger additional gain. The fused signal is mapped to positive token-level weights to reshape sequence-level advantages. Experiments on Qwen2.5-VL-3B/7B show that PEAR achieves the highest average performance among evaluated baselines on multimodal reasoning benchmarks, alongside leading results on textual reasoning and general multimodal benchmarks.

\section*{Limitations}
Although PEAR achieves performance gains on the evaluated tasks, this study has several limitations. Due to limited computational resources, we evaluate PEAR on Qwen2.5-VL-3B/7B but have not extended the experiments to larger models. We were also unable to conduct multiple independent runs with different random seeds for every experimental setting. Consequently, PEAR's performance at larger scales and the run-to-run variability of the reported results remain to be examined.

\bibliography{references}

\appendix

\section{Additional Training Details and Stabilization Ablation}

\subsection{Complete Training Configuration}

PEAR is implemented based on the publicly released PAPO codebase~\citep{papo}. ViRL39K~\citep{vl_rethinker} covers mathematics, science, charts, and social domains, with only questions and final answers provided. The training configurations for both model scales are summarized in Table~\ref{tab:training_configuration}.

\begin{table}[htbp]
    \centering
    \setlength{\tabcolsep}{6pt}
    \renewcommand{\arraystretch}{1.08}
    \resizebox{\columnwidth}{!}{%
    \begin{tabular}{@{}lcc@{}}
        \toprule
        \textbf{Configuration} & \textbf{3B} & \textbf{7B} \\
        \midrule
        Training data & ViRL39K & ViRL39K \\
        Training epochs & 2 & 2 \\
        Hardware & $8\times$ A100 & $8\times$ A100 \\
        Optimizer / precision & AdamW / BF16 & AdamW / BF16 \\
        Learning rate & $1\times10^{-6}$ & $1\times10^{-6}$ \\
        Weight decay & $1\times10^{-2}$ & $1\times10^{-2}$ \\
        Maximum gradient norm & 1.0 & 1.0 \\
        \makecell[l]{Global / rollout / mini-rollout\\batch size} & 128 / 384 / 128 & 128 / 384 / 128 \\
        Responses per prompt & 5 & 5 \\
        \makecell[l]{Maximum prompt / response\\length} & 4096 / 2048 & 4096 / 2048 \\
        Rollout temperature / top-$p$ & 1.0 / 0.99 & 1.0 / 0.99 \\
        Clipping range $(\epsilon_l,\epsilon_h)$ & $(0.2,0.28)$ & $(0.2,0.28)$ \\
        Dynamic-sampling range & $[0.01,0.99]$ & $[0.01,0.99]$ \\
        Entropy threshold $\tau_h$ & 0.5 & 0.5 \\
        Gate exponent & 2.0 & 2.0 \\
        Advantage modulation $\beta$ & 1.2 & 1.2 \\
        Exploration coefficient $\rho$ & Fixed at 1.0 & Cosine, $1.0\rightarrow0.1$ \\
        Entropy-penalty coefficient & 0 & 0.03 \\
        \bottomrule
    \end{tabular}%
    }
    \caption{Training configurations}
    \label{tab:training_configuration}
\end{table}

We use gradient checkpointing and fully sharded data parallel training. The rule-based reward combines answer accuracy and format compliance with weights of $0.9$ and $0.1$, respectively.

\paragraph{PEPO Baseline Results.}
For the 3B comparison, we use the Qwen2.5-VL-3B-Instruct results reported by PEPO on ViRL39K for benchmarks shared with our evaluation, marking unreported entries with dashes~\citep{pepo}. As PEPO does not report corresponding 7B results, we train Qwen2.5-VL-7B-Instruct using its publicly released code and evaluate it on all eight multimodal reasoning benchmarks under our evaluation protocol.

\subsection{Stabilization Designs for the 7B Model}
\label{app:stabilization_designs}

\paragraph{Cosine Annealing of the Exploration Gain.}
For Qwen2.5-VL-7B-Instruct, we introduce a step-dependent coefficient $\rho(s)$ on the exploration term and modify the fused signal as
\begin{equation}
    c_{i,t}^{(s)}
    = \hat{v}_{i,t}
    + \rho(s)\bigl(1-\hat{v}_{i,t}\bigr)e_{i,t}.
    \label{eq:7b_fusion}
\end{equation}
The coefficient follows a cosine schedule over training:
\begin{equation}
    \rho(s)
    = \rho_{\min}
    + \frac{\rho_{\max}-\rho_{\min}}{2}
    \left(1+\cos\left(\frac{\pi s}{S}\right)\right),
    \label{eq:rho_cosine}
\end{equation}
where $s$ denotes the current optimization step and $S=202$ is the total number of training steps. We set $\rho_{\max}=1.0$ and $\rho_{\min}=0.1$. The schedule assigns a larger exploration gain at the beginning of training, when broader policy exploration is needed, and gradually reduces its contribution as optimization proceeds. This prevents the exploration term from exerting an unnecessarily strong influence after the policy has begun to stabilize.

\paragraph{Entropy Penalty.}
Although token entropy is used by PEAR to estimate the exploration potential of different positions, it is not directly maximized as an optimization objective. Following the stabilization practice in PAPO~\citep{papo}, we add a lightweight regularizer for the 7B model. For a batch of $B$ sampled responses $\{\mathbf{y}_i\}_{i=1}^{B}$, we minimize the average negative log-probability over all valid response tokens conditioned on the original image:
\begin{equation}
    \begin{aligned}
    \mathcal{L}_{\mathrm{ent}}
    &= -\frac{1}{\sum_{i=1}^{B}|\mathbf{y}_i|}
    \sum_{i=1}^{B}\sum_{t=1}^{|\mathbf{y}_i|}\\
    &\quad
    \log \pi_{\theta}
    \left(y_{i,t} \mid \mathbf{x}_i,\mathbf{I}_i,\mathbf{y}_{i,<t}\right),
    \end{aligned}
    \label{eq:entropy_penalty}
\end{equation}
The entropy penalty is incorporated directly into the PEAR objective for 7B training:
\begin{equation}
    \mathcal{J}_{\mathrm{7B}}(\theta)
    = \mathcal{J}_{\mathrm{PEAR}}(\theta)
    - \lambda_{\mathrm{ent}}\mathcal{L}_{\mathrm{ent}},
    \qquad \lambda_{\mathrm{ent}}=0.03.
    \label{eq:7b_objective}
\end{equation}
This auxiliary regularizer empirically improves training stability, as examined in Appendix~\ref{app:stabilization_ablation}.

\subsection{Ablation of the Stabilization Designs}
\label{app:stabilization_ablation}

We conduct a progressive ablation on Qwen2.5-VL-7B-Instruct to evaluate the effects of the two stabilization designs. The first variant keeps the exploration-gain coefficient fixed at $\rho=1$ and does not use an entropy penalty, under which policy entropy can grow rapidly as training proceeds. The second introduces cosine annealing to gradually reduce the exploration gain and control this growth. Following the stabilization settings of related methods, the final configuration further adds a small entropy penalty as a regularizer and corresponds to the complete 7B model used in the main experiments. All variants otherwise share the training configuration described above.

\begin{figure}[htbp]
    \centering
    \includegraphics[width=\columnwidth]{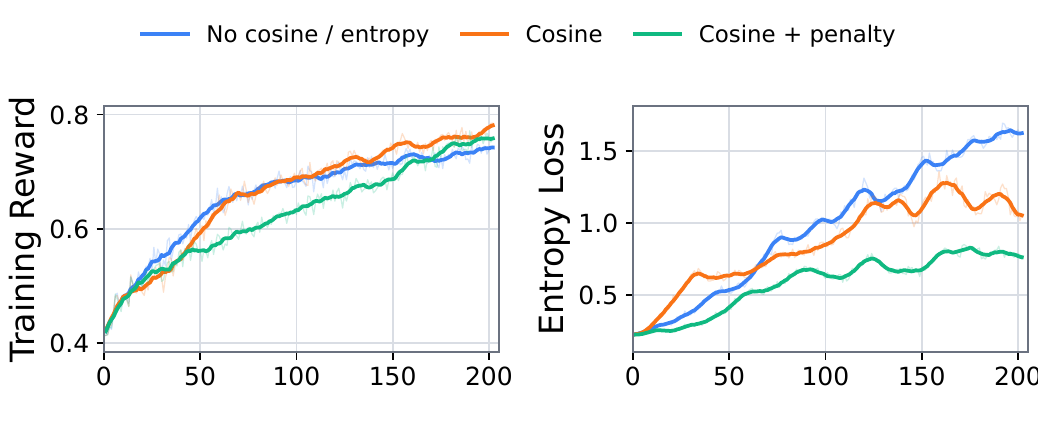}
    \caption{Training reward (left) and entropy loss (right) across optimization steps for the 7B stabilization settings in Table~\ref{tab:7b_stabilization_ablation}. Faint curves show raw values; solid curves show smoothed trends.}
    \label{fig:appendix_reward}
\end{figure}

As shown in Table~\ref{tab:7b_stabilization_ablation}, cosine annealing improves the average performance of the 7B model over the fixed-coefficient setting, and the entropy penalty provides a further gain.
\begin{table}[htbp]
    \centering
    \newcommand{\stabilizationcell}[2][1.12]{%
        \begingroup
        \sbox0{#2}%
        \makebox[\wd0][c]{\raisebox{0pt}[\ht0][\dp0]{\scalebox{#1}{\usebox0}}}%
        \endgroup
    }
    \setlength{\tabcolsep}{4.2pt}
    \renewcommand{\arraystretch}{1.08}

    \resizebox{\linewidth}{!}{
    \begin{tabular}{l|cccccc|c}
        \toprule
        \multicolumn{1}{c|}{\stabilizationcell[1.08]{\textbf{Setting}}}
        & \stabilizationcell[1.08]{\textbf{Geo3K}}
        & \stabilizationcell[1.08]{\textbf{We-Math}}
        & \stabilizationcell[1.08]{\textbf{MathVerse}}
        & \stabilizationcell[1.08]{\textbf{LogicVista}}
        & \stabilizationcell[1.08]{\textbf{MMMU-Pro}}
        & \stabilizationcell[1.08]{\textbf{MathVerse$_V$}}
        & \stabilizationcell[1.08]{\textbf{Avg.}} \\
        \midrule

        \stabilizationcell[1.08]{PEPO}
        & \stabilizationcell{39.62}
        & \stabilizationcell{68.82}
        & \stabilizationcell{66.11}
        & \stabilizationcell{46.20}
        & \stabilizationcell{35.33}
        & \stabilizationcell{62.17}
        & \stabilizationcell{53.04} \\

        \midrule

        \stabilizationcell[1.08]{PEAR ($\rho=1$)}
        & \stabilizationcell{42.45}
        & \stabilizationcell{68.45}
        & \stabilizationcell{68.65}
        & \stabilizationcell{46.42}
        & \stabilizationcell{36.35}
        & \stabilizationcell{65.42}
        & \stabilizationcell{54.62} \\

        \multicolumn{1}{r|}{\stabilizationcell[1.08]{+ Cosine Annealing}}
        & \stabilizationcell{43.24}
        & \stabilizationcell{\textbf{69.65}}
        & \stabilizationcell{70.01}
        & \stabilizationcell{47.37}
        & \stabilizationcell{\textbf{36.72}}
        & \stabilizationcell{65.99}
        & \stabilizationcell{55.50} \\

        \rowcolor{blue!6}
        \multicolumn{1}{r|}{\stabilizationcell[1.08]{+ Entropy Penalty}}
        & \stabilizationcell{\textbf{45.13}}
        & \stabilizationcell{\textbf{69.65}}
        & \stabilizationcell{\textbf{70.28}}
        & \stabilizationcell{\textbf{47.96}}
        & \stabilizationcell{36.55}
        & \stabilizationcell{\textbf{66.50}}
        & \stabilizationcell{\textbf{56.01}} \\

        \bottomrule
    \end{tabular}
    }
    \caption{Comparison with PEPO and ablation of PEAR's 7B stabilization designs. Best results are highlighted in bold.}
    \label{tab:7b_stabilization_ablation}
\end{table}

\section{Benchmark Details and Evaluation Protocols}
\label{app:benchmark_details}

\subsection{Multimodal Reasoning Benchmarks}

We evaluate multimodal reasoning on eight benchmarks spanning complementary forms of visual and logical reasoning. \textbf{Geo3K}~\citep{geo3k} evaluates geometry problem solving from diagrams and textual conditions. \textbf{DynaMath}~\citep{dynamath} measures the robustness of visual mathematical reasoning through dynamically varied problem instances. \textbf{We-Math}~\citep{we_math} evaluates fine-grained mathematical reasoning across hierarchically organized knowledge concepts. \textbf{MathVision}~\citep{mathvision} contains challenging multimodal mathematics problems collected from diverse competitions. \textbf{MathVerse}~\citep{mathverse} tests whether a model can ground mathematical reasoning in visual diagrams, while \textbf{MathVerse$_V$} denotes its vision-dependent subset. \textbf{LogicVista}~\citep{logicvista} focuses on logical reasoning grounded in visual contexts. \textbf{MMMU-Pro}~\citep{mmmu_pro} provides a more challenging expert-level evaluation across multiple academic disciplines.

For all eight benchmarks, we generate $k$ independent responses per example using a temperature of $1.0$, top-$p$ of $1.0$, a maximum generation length of 2048 tokens, and a maximum total sequence length of 9216 tokens. We extract the final boxed answer from each response and apply the corresponding benchmark evaluator. If $a_{j,r}\in\{0,1\}$ denotes the correctness of the $r$-th response to the $j$-th example, the Avg@$k$ score is
\begin{equation}
    \operatorname{Avg@}k
    = \frac{1}{N}\sum_{j=1}^{N}
      \frac{1}{k}\sum_{r=1}^{k} a_{j,r}.
    \label{eq:avgk}
\end{equation}
We set $k=8$ in these benchmarks. Thus, Avg@8 measures mean sampled-response accuracy rather than pass@8. For all models trained in our experiments, we use the same prompts, decoding parameters, answer extraction procedure, and evaluation scripts.

\subsection{Textual Reasoning Benchmarks}

The textual reasoning evaluation contains four benchmarks. \textbf{AIME24}~\citep{aime24} evaluates competition-level mathematical reasoning using problems from the 2024 American Invitational Mathematics Examination. \textbf{MATH-500}~\citep{math500} provides a representative subset of challenging problems from the MATH benchmark. \textbf{BBH}~\citep{bbh} aggregates difficult tasks that require multi-step symbolic, algorithmic, and logical reasoning. \textbf{MMLU-Pro}~\citep{mmlu_pro} evaluates robust knowledge-intensive reasoning with more challenging questions and expanded answer choices.

For AIME24, we sample 16 responses for each problem and report Avg@16, defined analogously to Eq.~(\ref{eq:avgk}). For MATH-500, BBH, and MMLU-Pro, we generate one response per example and report Avg@1 accuracy. The detailed decoding settings are summarized in Table~\ref{tab:textual_eval_settings}.

\begin{table}[htbp]
    \centering
    \setlength{\tabcolsep}{5pt}
    \renewcommand{\arraystretch}{1.08}
    \resizebox{\columnwidth}{!}{%
    \begin{tabular}{@{}lcccc@{}}
        \toprule
        \textbf{Benchmark} & \textbf{Responses} & \makecell{\textbf{Temp. / top-$p$}\textbf{/ top-$k$}} & \makecell{\textbf{Model }\textbf{length}} & \makecell{\textbf{Output }\textbf{length}} \\
        \midrule
        MATH-500 & 1  & $1.0/0.95/50$ & 16,384 & 16,384 \\
        AIME24   & 16 & $1.0/0.95/50$ & 16,384 & 16,384 \\
        BBH      & 1  & $0/1.0/-1$    & 8,192  & 2,048 \\
        MMLU-Pro & 1  & $0/1.0/-1$    & 8,192  & 2,048 \\
        \bottomrule
    \end{tabular}%
    }
    \caption{Decoding settings for textual reasoning evaluation.}
    \label{tab:textual_eval_settings}
\end{table}

All textual benchmarks are evaluated in text-only mode using vLLM with seed 42 and an ARES-style prompt that places the reasoning process in \texttt{<think>} tags and the final answer in \texttt{\textbackslash boxed\{\}}. We extract the last boxed or framed expression when available and otherwise use final-answer phrases or the last nonempty line. MATH-500 and AIME24 use normalized exact or numerical matching, BBH uses normalized text or option matching, and MMLU-Pro matches the extracted option letter.

\subsection{General Multimodal Benchmarks}

We evaluate general multimodal capability on \textbf{MMStar}~\citep{mmstar}, \textbf{MMMU}~\citep{mmmu}, \textbf{MME}~\citep{mme}, and \textbf{POPE}~\citep{pope}. MMStar emphasizes samples that require genuine visual content to answer; MMMU covers expert-level multimodal understanding across a broad range of disciplines; MME evaluates both perceptual and cognition-oriented capabilities; and POPE measures object hallucination in visual question answering.

We conduct inference, prediction post-processing, and metric computation for these benchmarks using VLMEvalKit~\citep{vlmevalkit}, and report Avg@1 under the corresponding official evaluation protocols. Accuracy is reported for MMStar and MMMU. Following the presentation in Table~\ref{tab:general_capability}, the original MME score is divided by its maximum value of 2800 and reported as a percentage, while POPE is reported using the F1 score. During the 7B evaluation on MMMU, we observe that the default rule-based evaluator cannot extract a valid answer from a small subset of responses because of their output format. We use GPT-4o~\citep{gpt} only for these parsing-failure cases to judge whether the generated response is correct; all responses that can be parsed normally remain scored by the standard VLMEvalKit evaluator. This fallback rule is applied uniformly to all 7B outputs that satisfy the same parsing-failure condition. The common generation settings are summarized in Table~\ref{tab:general_eval_settings}.

\begin{table}[htbp]
    \centering
    \setlength{\tabcolsep}{6pt}
    \renewcommand{\arraystretch}{1.08}
    \resizebox{\columnwidth}{!}{%
    \begin{tabular}{@{}cc@{}}
        \toprule
        \textbf{Setting} & \textbf{Value} \\
        \midrule
        Responses per example & 1 \\
        Temperature / top-$p$ & $0/1.0$ \\
        Maximum model / output length & $32{,}768/2{,}048$ \\
        Minimum / maximum image pixels & $1{,}003{,}520/12{,}845{,}056$ \\
        Prompt and evaluator & Dataset-specific VLMEvalKit defaults \\
        \bottomrule
    \end{tabular}%
    }
    \caption{Common settings for general multimodal evaluation.}
    \label{tab:general_eval_settings}
\end{table}

\section{Word-Cloud Construction Details}
\label{app:wordcloud_construction}

Following the entropy-based token visualization in ARES~\citep{ares}, we construct the word cloud from the selected token subset rather than all response tokens. For the 4,078 correct trajectories generated on MathVerse and used in Fig.~\ref{fig:wordcloud}, we follow the definitions in the main text by first applying the logarithmic transformation to visual dependency and then independently normalizing visual dependency and token entropy within each trajectory. We retain tokens with normalized visual dependency no greater than $0.1$ and normalized entropy greater than $0.5$. The selected token positions are merged into complete English words and converted to lowercase; English stop words, punctuation-only entries, formatting tokens, and \LaTeX{} commands are then removed. This produces 91,680 word occurrences and 4,682 unique words, with word size in Fig.~\ref{fig:wordcloud} proportional to occurrence frequency. These operations are used only to construct the visualization.

\end{document}